%% file: acl_latex.tex
\pdfoutput=1

\documentclass[11pt]{article}

\usepackage[final]{acl}

\usepackage{times}
\usepackage{latexsym}
\usepackage{amsmath, amssymb}
\usepackage{graphicx} 
\usepackage{float} 
\usepackage{subfigure} 
\usepackage{makecell}
\usepackage{multirow}
\usepackage{graphicx}
\usepackage{tabularx}
\usepackage{color, soul}
\usepackage{colortbl}
\usepackage{tcolorbox}
\usepackage{color,xcolor}
\usepackage{booktabs}
\usepackage{wrapfig}

\usepackage[T1]{fontenc}

\usepackage[utf8]{inputenc}

\usepackage{microtype}

\usepackage{inconsolata}
\usepackage{pifont}
\usepackage{enumitem}
\let\oldding\ding
\renewcommand{\ding}[2][1]{\scalebox{#1}{\oldding{#2}}}

\title{WKVQuant: Quantizing Weight and Key/Value Cache for Large Language Models Gains More}

\author{Yuxuan Yue\thanks{Equal contribution.} \\ Harbin Institute of Technology (Shenzhen) \And
        Zhihang Yuan$^*$ \\ Houmo AI \AND
        Haojie Duanmu \\ Shanghai Artificial Intelligence Laboratory \\ Shanghai Jiao Tong University \And
        Sifan Zhou \\ Houmo AI \And
        Jianlong Wu \and Liqiang Nie \\
        Harbin Institute of Technology (Shenzhen)}

\begin{document}
\maketitle

\input{sec/0_abstract}
\input{sec/1_intro}
\input{sec/2_method}
\input{sec/3_experiments}

\newpage


\section*{Limitations}
Large Language Models (LLMs) face significant problem when deploying because of the vast size. In this work, we analyze the memory consumption of different parts of LLMs (i.e., weights, Key/Value cache, and other temporary activations) and their corresponding quantization difficulties. From the analysis we know that the temporary activations are of low value to be quantized. We thus propose \emph{WKVQuant}, which optimizes the exclusive quantization of weight and KV cache. Although this approach can both maintain model accuracy and reduce memory consumption, it is limited by not quantizing temporary activations. 

Specifically, (1) this limitation means increased memory consumption since the temporary activations are remained in full-precision, especially when in prefill stage and when the batchsize/sequence length is large. We don't choose to quantize this part of activations because it consumes few memory space in most cases. 
(2) This limitation means that our approach can not utilize faster computation units, such as Nvidia's INT8 acceleration units. Weight-activation methods quantized all operation variables into the same bit-width, thus having a good implementation on the mentioned units.
Although there should be a time cost difference, we claim that the gap is minimal because the inference of LLMs is primarily constrained by memory access. 

In addition, we note that the cross-block reconstruction regularization will lead to an increase in the time required for optimizing quantization parameters. Specifically, for a 7b model, the process takes approximately 3 hours, while for a 13b model, it takes around 4 hours. However, it is worth mentioning that the process is typically performed only once before deployment. Therefore, we believe that the additional time is generally acceptable in the majority of cases.

\bibliography{custom}
\clearpage
\appendix
\input{sec/4_appendix}

\end{document}

%% file: sec/0_abstract.tex
\begin{abstract}
\label{sec:abs}
 Large Language Models (LLMs) face significant deployment challenges due to their substantial memory requirements and the computational demands of auto-regressive text generation process. 
 This paper addresses these challenges by focusing on the quantization of LLMs, a technique that reduces memory consumption by converting model parameters and activations into low-bit integers. 
 We critically analyze the existing quantization approaches, identifying their limitations in balancing the accuracy and efficiency of the quantized LLMs. 
 To advance beyond these limitations, we propose \emph{WKVQuant}, a PTQ framework especially designed for quantizing weights and the key/value (KV) cache of LLMs. 
 Specifically, we incorporates past-only quantization to improve the computation of attention. Additionally, we introduce two-dimensional quantization strategy to handle the distribution of KV cache, along with a cross-block reconstruction regularization for parameter optimization.
 Experiments show that \emph{WKVQuant} achieves almost comparable memory savings to weight-activation quantization, while also approaching the performance of weight-only quantization.
\end{abstract}

%% file: sec/1_intro.tex
\section{Introduction}
\label{sec:intro}
\vspace{-2mm}



Large language models (LLMs) such as GPT \cite{brown2020language,ouyang2022training}, OPT \cite{zhang2022opt}, and LLaMA \cite{touvron2023llama,touvron2023llama2} are essential in natural language processing, demonstrating unparalleled abilities to understand and generate text.
However, their large size poses significant challenges for deployment.
Firstly, the large number of weights in LLMs consumes a considerable amount of memory. For instance, the LLaMA-13b model requires approximately 26GB of memory when stored in FP16 format, which can only be accommodated by high-end GPUs.
Secondly, due to the auto-regressive nature of text generation in LLMs, it is common practice to store certain intermediate results for reuse during iteration to avoid redundant computations. 
These stored data, known as key/value cache (KV cache), occupy memory that increases as the batch size and sequence length grow~\cite{miao2023towards}. 
\begin{figure}[!tbp]
        \centering
        \includegraphics[]{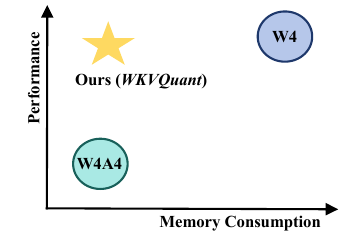} 
        \caption{Memory-Performance curve.}
        \label{figure:curve}
        \vspace{-5mm}
\end{figure}
To address the issue of excessive memory usage in LLMs, quantization has emerged as a widely adopted solution~\cite{zhao2023survey}. By quantizing tensors into low-bit integers to represent numerical values, the memory footprint can be drastically reduced by 2-8 times. Existing quantization methods for LLMs can be categorized into two types, including weight-only quantization and weight-activation quantization~\cite{zhao2023survey}. As the KV cache is not quantized, weight-only quantization methods have a minimal impact on model accuracy, but is not effective enough at reducing memory consumption. In contrast, the weight-activation methods quantize both the weights and activations, including the KV cache. However, it suffers from a significant drop in accuracy. Both of these quantization paradigms have their merits and drawbacks, making it infeasible to simultaneously achieve the benefits of both approaches. 

\begin{figure}[!tbp] 
    \centering
    \includegraphics[width=\linewidth]{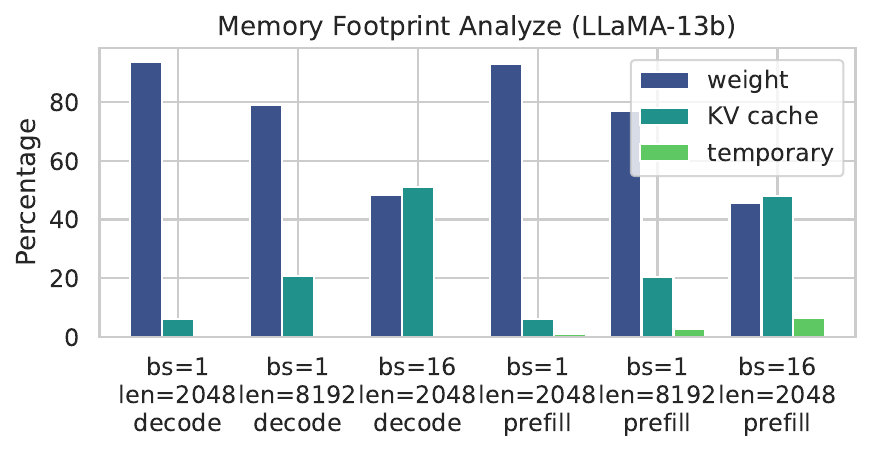} 
    \caption{The memory usage proportions of weight, KV cache, and temporary storage on LLaMA-2-13b.}
    \label{im_memory_footprint}
\end{figure}

To overcome the dilemma between the accuracy and efficiency of the quantized LLMs, we consider the activation quantization on a finer-grained scale. The activations in LLMs include KV cache and other temporary activations. In this paper, we demonstrate that it is not cost-effective to quantize temporary activations.
Specifically, 
\raisebox{-0.5pt}{\ding[1.1]{182\relax}}
our analysis reveals that the temporary activation of memory is short-lived, as the memory content generated by one Transformer layer can be completely overwritten by the next layer.
The memory footprint of temporary activations is significantly smaller compared to the memory usage of weights and the KV cache as shown in Figure~\ref{im_memory_footprint}.
\raisebox{-0.5pt}{\ding[1.1]{183\relax}}
Temporary activations is highly sensitive to quantization, leading to high accuracy drop~\cite{dettmers2022llm,xiao2022smoothquant}. 
\raisebox{-0.5pt}{\ding[1.1]{184\relax}}
In decoding phase of LLM, as the computation is bound by memory access of weights and KV cache, quantizing temporary activations to utilize low-precision arithmetic units does not significantly reduce inference latency. Therefore, exclusive quantization of weights and KV cache is a beneficial trade-off between accuracy and memory savings. 
Existing methods are not designed for the exclusive quantization of weights and KV cache, which can lead to a decrease in accuracy. In this paper, we propose the \emph{WKVQuant} framework, which is specifically designed for quantizing weights and KV cache.
\raisebox{-0.5pt}{\ding[1.1]{182\relax}} We introduce a new approach called Past Only Quantization (POQ) to enable higher precision in the Attention mechanism. Instead of discarding the full-precision Key and Value values (KV) after quantization, we temporarily retain them. This allows us to use unquantized KV during matrix multiplication in the Attention mechanism, improving the accuracy of Attention computation. 
In the prefill phase, POQ allows the network to achieve the same level of accuracy as weight-only quantization.
In the decode phase, POQ combines the current unquantized KV with the previously quantized KV cache for prediction, improving the prediction accuracy.
\raisebox{-0.5pt}{\ding[1.1]{183\relax}} We observe significant variations in numerical values between channels and tokens in KV cache. Towards this issue, we propose Two-dimensional Quantization (2D-Quantization).
Static channel smoothing aligns large values between channels, while dynamic token-wise quantization addresses variations between tokens.
\raisebox{-0.5pt}{\ding[1.1]{184\relax}} We discover a bias in the reconstruction loss function used for parameter optimization in the existing PTQ method~\cite{2023omniquant}. So we further introduce cross-block reconstruction regularization to reduce such bias and obtain better quantized parameters and smoothing parameters, thereby reducing quantization errors.

We evaluate the performance of \emph{WKVQuant} on the LLaMA (Touvron et al., 2023) and LLaMA-2 (Touvron et al., 2023) models. Our experiments demonstrate that the W4KV4 quantized network with \emph{WKVQuant} achieves significantly higher accuracy compared to the W4A4 quantized network. Specifically, when generating short sequences, W4KV4 performs equally well as the W4 quantization. For long sequence generation, W4KV4 exhibits performance close to that of W4 quantization. Meanwhile, the memory consumption of W4KV4 is nearly identical to that of W4A4. 

In conclusion, our proposed \emph{WKVQuant} provides a promising trade-off between accuracy and efficiency as shown in Figure~\ref{figure:curve}.



\section{Related Work}
\label{sec:Related Work}
\vspace{-2mm}
Existing quantization methods for LLMs can be classified into two types: weight-only quantization and weight-activation quantization~\cite{zhao2023survey}.

\vspace{-1mm}
\paragraph{Weight-only Quantization for LLMs.} For weight-only quantization, some works make efforts in the realm of Quantization-Aware Training (QAT), LLM-QAT\cite{liu2023llm} innovatively tackles the challenges in acquiring training data for LLMs by leveraging pre-trained models for data-free distillation. Moreover, works such as QLORA~\citep{dettmers2024qlora}, PEQA~\cite{kim2023peqa}, QA-lora~\cite{xu2023qa} and LoftQ~\cite{li2023loftq} leverage Parameter Efficient Fine-Tuning (PEFT) techniques on performing fine-tuning tasks while achieving model compression. For Post-Training Quantization (PTQ) on LLMs\cite{kim2023squeezellm, dettmers2023spqr, shang2023pb}, GPTQ~\cite{frantar2022gptq} and QuIP~\cite{chee2023quip} achieve high compression rates by optimizing matrix multiplications operation and propose a novel layer-wise quantization solution. Besides, AWQ~\cite{lin2023awq} and OWQ~\cite{lee2023owq} take into account the impact of activation outliers on weight quantization, showing quantization performance improvements. PB-LLM ~\cite{shang2023pb} exploits the salient-weight
property of LLM and achieves extreme quantization to the lowest possible bit. While weight-only quantization can alleviate the computational burdens, its impact on memory usage and acceleration is still limited compared to weight-activation quantization.



%

\vspace{-1mm}
\paragraph{Weight-activation Quantization for LLMs.} Different with weight-only quantization methods, the weight-activation quantization methods~\cite{wei2022outlier,wei2023outlier,li2023norm, yuan2023asvd} quantize both the weights and activations, including the KV cache. The most challenge for quantizing activations lies in outliers, which often lead to great quantization error. To tackle this issue, ZeroQuant~\cite{yao2022zeroquant} proposes a fine-grained hardware-friendly quantization scheme for both weight and activations. SmoothQuant~\cite{xiao2022smoothquant} migrates the quantization difficulty from activations to weights with a mathematically equivalent transformation (i.e., per-channel scaling).  OmniQuant~\cite{2023omniquant} further enhances performance by training the quantization parameters. RPTQ~\cite{yuan2023rptq} proposes grouped quantization after clustering similar channels to tackle the outliers. While these methods have mitigated the quantization error, they commonly focus on addressing outliers in intermediate computational results. In this study, our principal emphasis is on quantizing weights and KV cache with PTQ techniques. Particularly, to our knowledge, \emph{WKVQuant} is the first method for tackling weight and KV cache quantization predicaments.

%% file: sec/2_method.tex
\section{Method}
\begin{figure}[!tbp] 
    \centering
    \includegraphics[width=\linewidth]{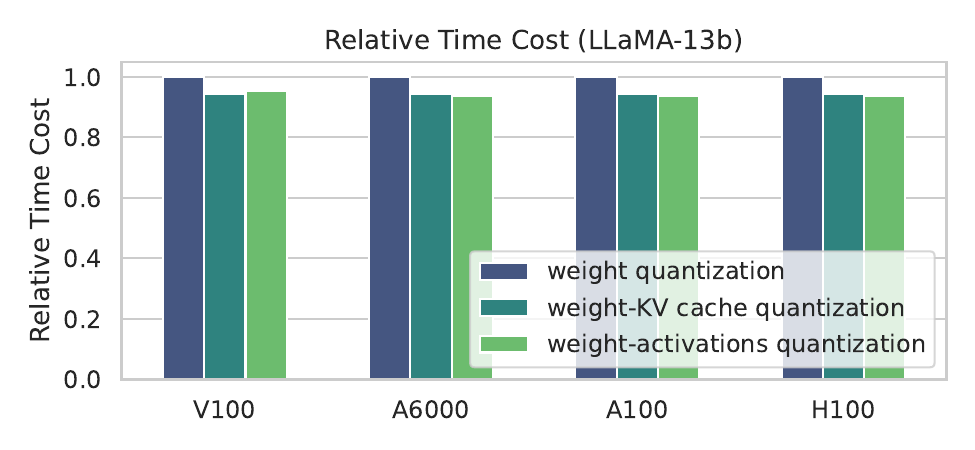} 
    \caption{Relative time cost to generate 2048 tokens for a 2048 length input prompt using different quantization settings on LLaMA-2-13b (batchsize=1).}
    \vspace{-2mm}
    \label{im_time}
\end{figure}
In this section, we first analysis the memory consumption when deploying LLMs in Section~\ref{sec:Memory Consumption Analysis} and present the proposed \emph{WKVQuant} in Section~\ref{sec:Weight-KV Cache Quantization}. We then describe the details of \emph{WKVQuant} in the following sections.







\vspace{-2mm}
\subsection{Memory Consumption Analysis}\label{sec:Memory Consumption Analysis}
Deploying LLMs faces a significant challenge due to their enormous memory consumption.
The memory usage of LLMs is mainly contributed by three resources:  \raisebox{-0.5pt}{\ding[1.1]{182\relax}} the weights of LLMs need to be stored in memory. For example, LLaMA-13b with 13 billion weights occupies around 26GB of memory in FP16 format. \raisebox{-0.5pt}{\ding[1.1]{183\relax}} Temporary activations are generated during inference. For instance, Transformer inputs are kept in memory until the residual connection is executed. \raisebox{-0.5pt}{\ding[1.1]{184\relax}} For auto-regressive LLMs, caching key and value activations (KV cache) into memory is necessary for subsequent token generation.

We use LLMViewer~\cite{hahnyuanLLMViewer} to analyze memory consumption and show the results in Figure~\ref{im_memory_footprint}. 
The analysis reveals that the memory footprint of temporary activations is relatively small, especially in decode phase. This is due to their short lifespan, as their memory can be released after usage.
In contrast, the memory allocated for the KV cache cannot be released until the completion of a full answer generation process, which involves multiple inference passes through the network. Moreover, the memory consumption of the KV cache increases with larger batch sizes and longer input sequences as the model needs to store more KV pairs.

\vspace{-2mm}
\subsection{Weight-KV Cache Quantization}
\label{sec:Weight-KV Cache Quantization}
\begin{table}[!tbp]
\centering
\begin{tabular}{@{}lccc@{}}
\toprule
\normalsize{~}& \small{Weight} & \small{KV Cache} & \small{Temporary Act} \\ \midrule
\small{INT8} &       \small{4.89}       &        \small{4.88}        &       \small{4.92}          \\
\small{INT4} &        \small{4.99}      &        \small{5.27 }       &       \small{785.56}        \\ \bottomrule
\end{tabular}
\caption{Perplexity on WikiText2 under quantizing each part of LLaMA-2-13b (perplexity of FP16 model is 4.88). Weight is quantized by GPTQ, and activations are dynamically per-token quantized. Act denotes activations.} 
\label{tab_sensitivity} 
\vspace{-3mm}
\end{table}

As mentioned in Section \ref{sec:Related Work}, there are two primary quantization methods for LLMs: weight-only quantization and weight-activation quantization. Weight-only quantization can compress the model weights to 4 bits or lower while minimizing the impact on accuracy. However, these method still result in relatively higher memory usage due to lack of activation quantization. Differently, weight-activation quantization aims to minimize the memory usage of LLMs to the greatest extent possible. However, quantizing activations has a larger impact on model accuracy compared to weight-only quantization. Furthermore, we opt to quantize the representative LLaMA-2-13b model and evaluate perplexity on the WikiText2~\cite{merity2016pointer} dataset to investigate the impact of each part. As shown in Table~\ref{tab_sensitivity}, we observe that both 4-bit quantization of KV cache and temporary activations lead to decreased network performance, with temporary activations having a particularly significant drop.

\begin{figure}[!tbp] 
    \centering
    \includegraphics[width=\linewidth]{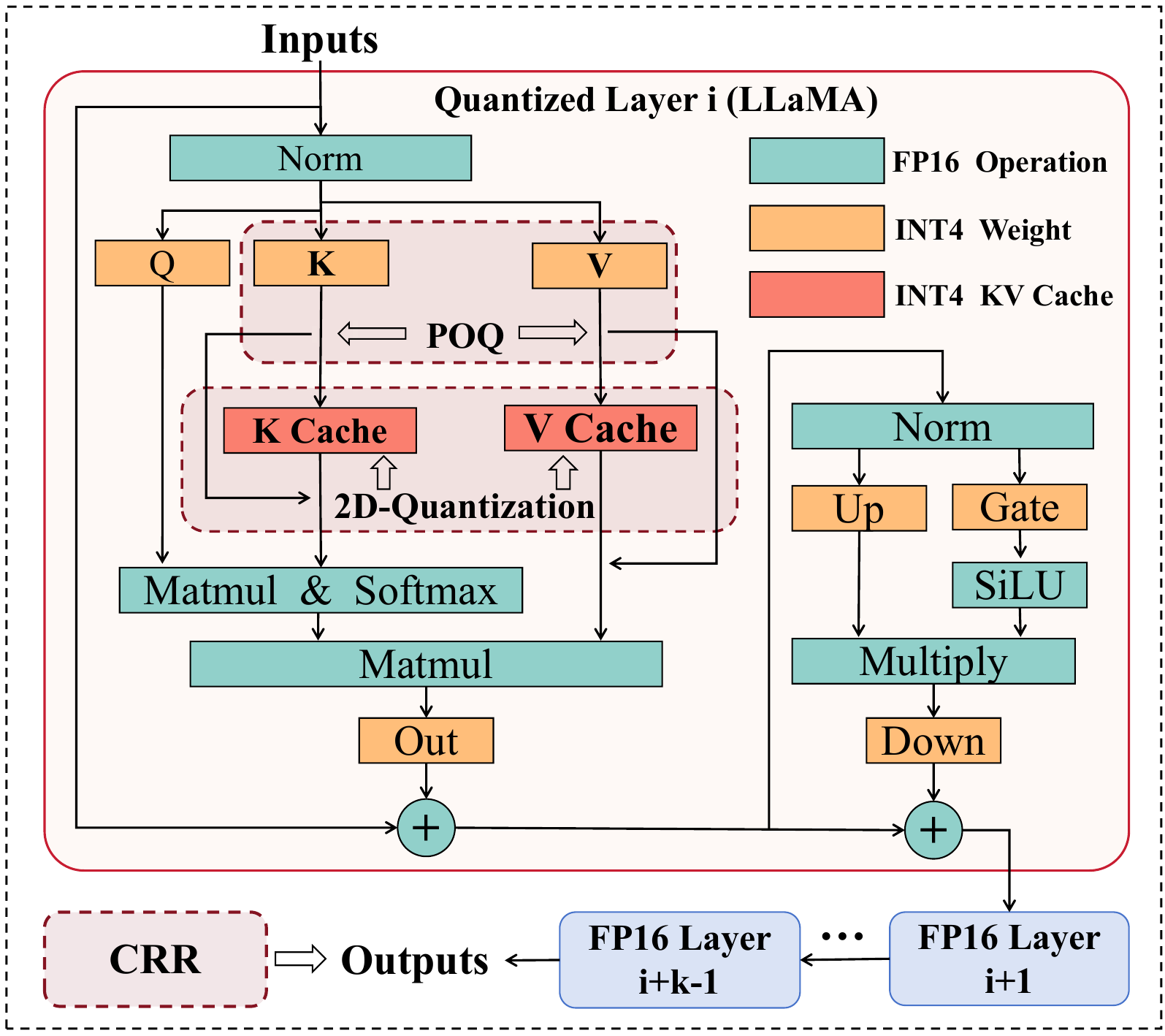}
    \caption{The framework of \emph{WKVQuant}.
    }
    \label{figure:overview}
\vspace{-5mm}
\end{figure}
Based on above observations, we believe that exclusively quantizing the weights and KV cache is a favorable choice. However, not quantizing temporary activations has two drawbacks: increased memory usage and the inability to utilize faster computation units, such as Nvidia's INT8 acceleration units. Nevertheless, these drawbacks do not significantly impact network inference. Firstly, as discussed earlier, temporary activations occupy minimal memory. Secondly, the unavailability of faster computation units does not greatly affect the inference speed. Figure~\ref{im_time} shows that the time cost of weight-KV cache quantization is nearly the same as weight-activation quantization. This is because the inference of LLMs is primarily constrained by memory access. Therefore, exclusive quantization of weights and KV cache is a beneficial trade-off between retaining model accuracy and achieving memory savings when compared to quantizing both weights and activations. 

To achieve this trade-off, we propose a Post-Training Quantization (PTQ) framework, \emph{WKVQuant}, which is specifically tailored for quantizing weights and KV cache of LLMs. Notably, to our knowledge, WKVQuant is the first method developed to address the exclusive quantization of weights and KV cache. The overview of \emph{WKVQuant} is shown in Figure~\ref{figure:overview}. Our \emph{WKVQuant} comprises three main features: Past-Only Quantization (POQ) to enhance attention computation, 2D-Quantization strategy to handle the distribution of KV cache, and cross-block reconstruction regularization for parameter optimization. We will describe the details in the following section.
\begin{figure}[!tbp] 
    \centering
    \includegraphics[width=0.95\linewidth]{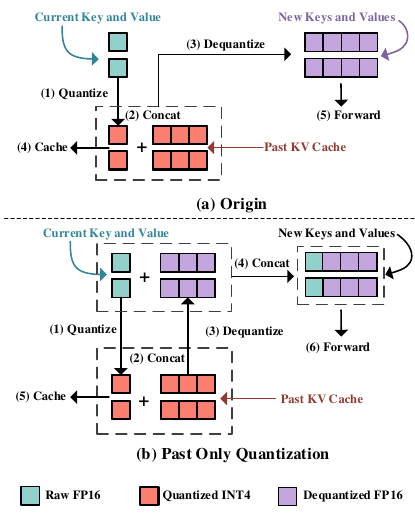} 
    \caption{Comparison between original quantization method and Past Only Quantization (POQ) for KV cache. POQ utilizes the current key and value in raw full-precision and only quantizes the past KV cache.}
    \label{figure:Past Only Quantization}
\vspace{-3mm}
\end{figure}
\subsection{Past Only Quantization}
\label{sec:Past Only Quantization}
\vspace{-1mm}
In auto-regressive token generation, it is a common practice to store the keys and values of each layer into cache. This KV cache serves as input for the next token generation process, reducing redundant computations between steps. While this computation optimization is beneficial for LLMs, it inevitably results in a higher memory footprint, especially when dealing with large batch size or long input sequence  (see Figure \ref{im_memory_footprint}).
Furthermore, as shown in Figure~\ref{figure:Past Only Quantization}. (a), we can observe that in the existing quantization method, the current key and value undergoes quantization and de-quantization before being passed into inference, which is a suboptimal approach. 


To this end, we propose a novel inference process called Past Only Quantization (POQ) to preserve the accuracy of the current key and value. As shown in Figure~\ref{figure:Past Only Quantization}.(b), POQ does not quantize the representations generated from the current token. Instead, it uses the original full-precision representations when merging them with the de-quantized past KV cache during the forward process. 
Only the past KV loaded from KV cache are quantized in current decoding inference phase.
By using unquantized KV during matrix multiplication in the Attention mechanism, POQ enhances the accuracy of Attention computation.
During the prefill phase, POQ enables the network to attain the same level of accuracy as weight-only quantization since all the KV used in Attention are full-precision.
In the decode phase, POQ combines the current unquantized KV with the previously quantized KV cache for prediction, thereby improving the prediction accuracy.

\subsection{Two-dimensional Quantization}\label{sec:Two-dimensional Quantization}

Previous studies have highlighted significant variations in numerical values among activation channels~\cite{xiao2022smoothquant,wei2023outlier}. Similarly, we have observed substantial variations between channels and tokens in the KV cache. Quantizing them directly using the same parameters leads to substantial quantization errors.
To tackle this issue, we introduce Two-dimensional Quantization. This approach employs static channel smoothing to align large values across channels and dynamic token-wise quantization to address variations between tokens.



The projection process of attention operation takes an input token sequence $\textbf{X} \in \mathbb{R}^{T \times {C_{\text{in}}}}$, a weight matrix $\textbf{W} \in \mathbb{R}^{{C_{\text{in}}} \times {C_{\text{out}}}}$, and a bias vector $\textbf{B} \in \mathbb{R}^{1\times{C_{\text{out}}}}$, where $T$, ${C_{\text{in}}}$, and ${C_{\text{out}}}$ represent token length, input channels, and output channels, respectively. The mathematically projection process of attention operation can be expressed as: $\textbf{Y} = \textbf{XW} + \textbf{B}$, where \textbf{Y} represents KV cache in our settings. 

Inspired by previous methods~\cite{xiao2022smoothquant,wei2023outlier}, we introduce a learnable shifting parameter $\delta \in \mathbb{R}^{1\times{C_{\text{out}}}}$ to align the centers of each channel, and also a learnable smoothing parameter $s \in \mathbb{R}^{1\times{C_{\text{out}}}}$ to adjust each channel to appropriate range. By introducing those parameters, the projection process of attention operation can be transformed into:

\vspace{-2mm}
\begin{small}
\begin{equation}
\vspace{-2mm}
\textbf{Y} = \underbrace{(\textbf{Y}-\delta)\oslash s}_{\Tilde{\textbf{Y}}} \odot s + \delta,
\label{eq:transformed}
\end{equation}
\end{small}
\noindent where\textquoteleft$\oslash$\textquoteright and \textquoteleft$\odot$\textquoteright represent element-wise division and multiplication, respectively. $\Tilde{\textbf{Y}}$ is the smoothed key or value after channel-wise shifting and scaling. The value difference among channels are significantly decreased by smoothing. Note that part of the parameters can be absorbed into the original linear weight and bias, which can be formulated as:

\begin{small}
\vspace{-2mm}
\begin{equation}
\vspace{-1mm}
\Tilde{\textbf{Y}}=(\textbf{XW}+\textbf{B}-\delta) \oslash s
=\textbf{X}\underbrace{\textbf{W}\oslash s}_{\Tilde{\textbf{W}}} + \underbrace{(\textbf{B}-\delta)\oslash s}_{\Tilde{\textbf{B}}},
\label{eq:absorbed}
\end{equation}
\end{small}
\noindent where $\Tilde{\textbf{W}}$ and $\Tilde{\textbf{B}}$ denote transformed weight and bias, respectively. After that, the quantization operation is applied to the smoothed weight and bias.

\vspace{+2mm}

To further suppress the outliers in the token dimension, we utilize dynamic token-wise fine-grained quantization as formulated by:
\begin{small}
\vspace{-2mm}
\begin{equation}
\vspace{-1mm}
\label{eq:shifted-symmetric}
    \begin{aligned}
&Q_t(\Tilde{\textbf{Y}})={\rm clamp}(\lfloor\frac{\Tilde{\textbf{Y}}-m}{n}\rceil, -2^{N-1}, 2^{N-1}-1),\\
&{\rm where}\ n=\frac{{\rm max}({\rm abs}(\Tilde{\textbf{Y}}-m))}{2^{N-1}},\ m={\rm mean}(\Tilde{\textbf{Y}}),
    \end{aligned}
\end{equation}
\end{small}
\noindent where $Q_t$ represents quantization process. $\lfloor\cdot\rceil$ denotes round operation and $N$ is the target bit number. Note that the ${\rm max}$, ${\rm abs}$, and ${\rm mean}$ operations are dynamically calculated in the token dimension. By shifting each value of every token close to their average, the outliers have less effect on per-token quantization. In addition, the fine-grained quantization method can be applied to this step, which calculates the $m$ and $n$ in a finer-grained (such as 128 values as a group). This approach can further enhance the quantization performance. 


For the weight quantization, we adopt the method in Omniquant~\cite{2023omniquant}, which is formulated as:

\begin{small}
    \begin{equation}
        \textbf{W}_q={\rm clamp}\left(\lfloor\frac{\textbf{W}}{h}\rceil+z,0, 2^{N-1}\right),
    \end{equation}\label{eq:weight quantization}
\end{small}
where $h=(\gamma{\rm max}(\textbf{W})-\beta{\rm min}(\textbf{W}))/2^{N-1}$, $z=-\lfloor\beta{\rm min}(\textbf{W})/h\rceil$. $\gamma$ and $\beta$ are trainable clipping parameters. 
\begin{figure}[!tbp] 
    \centering
    \includegraphics[width=\linewidth]{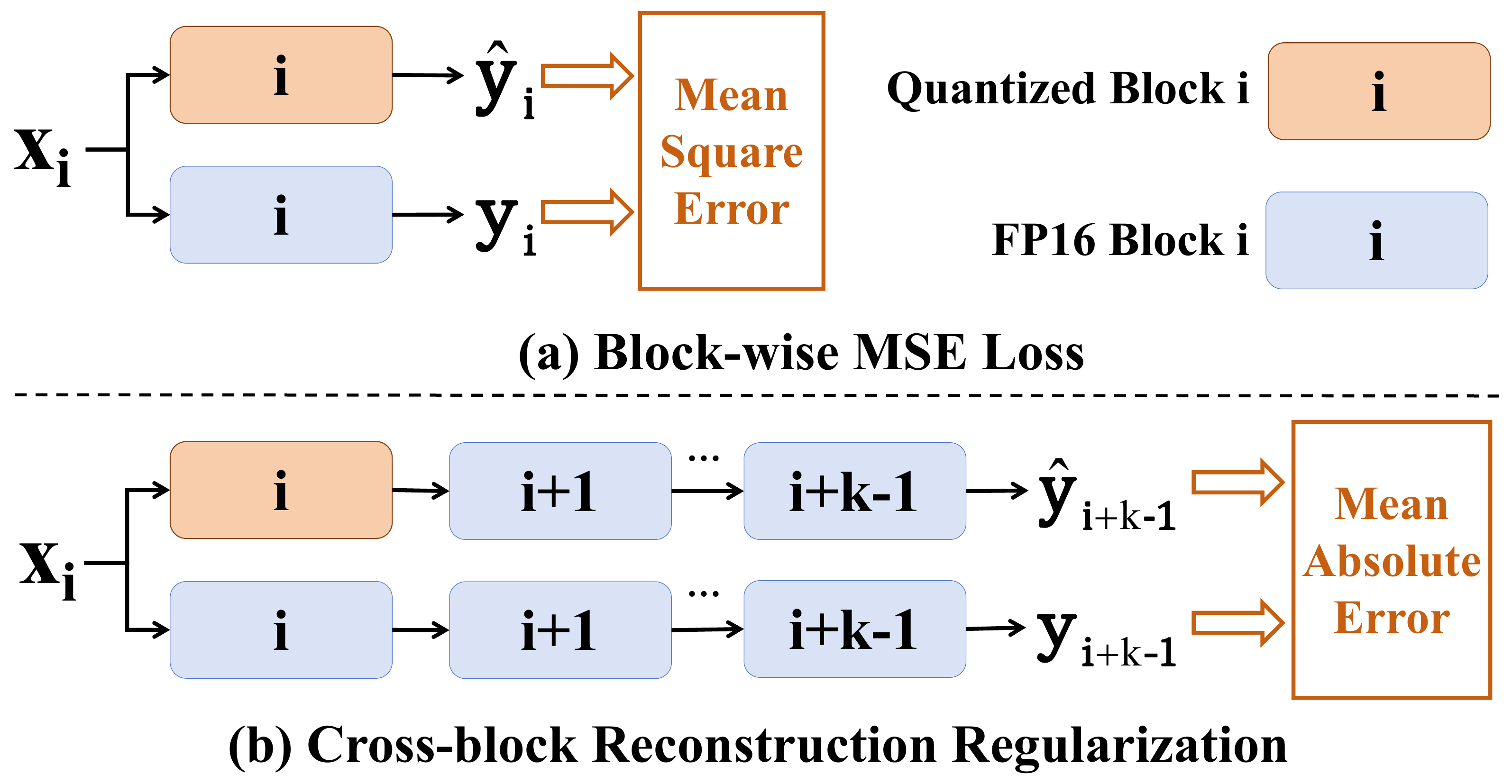} 
    \caption{Comparison between Block-wise Mean Square Error (MSE) loss and Cross-block Reconstruction Regularization (CRR): CRR computes Mean Absolute Error (MAE) in the block dimension of $i+k-1$ while Block-wise MSE loss computes MSE in the block dimension of $i$.}
    \label{figure:Cross-block Reconstruction Regularization}
\vspace{-3mm}
\end{figure}
\subsection{Cross-block Reconstruction Regularization}\label{sec:Cross-block Reconstruction Regularization}

Based on above description, there are some parameters need to be determined, including the clipping parameters $\gamma$ and $\beta$, as well as the smoothing parameters $s$ and $\delta$. Previous methods~\cite{2023omniquant} optimized these parameters using gradient descent. The approach involves calculating the local reconstruction loss between the output of the quantized Transformer layer and the full-precision Transformer layer, computing the gradients of these parameters with respect to the loss, and updating the parameters accordingly. However, it is worth noting that the local reconstruction loss introduces a bias and does not align with the final task loss~\cite{ptq4vit,Liu_2023_CVPR, zhou2024lidar}.
This is due to the varying impact of each activation in the Transformer layer on the final network output.
Additionally, it has been observed that outliers within the activations significantly affect on the Mean Squared Error (MSE) loss, which is square-based.


We propose a method called Cross-block Reconstruction Regularization (CRR) to mitigate this issue without significantly increasing computational and memory overhead. As shown in Figure~\ref{figure:Cross-block Reconstruction Regularization}, we pass the output of the Transformer layer to the subsequent layers, and then compute the loss by calculating the difference between the output results of the subsequent layers.
Compared to the block-wise MSE loss, CRR provides a closer approximation to the error of network's final output, resulting in a smaller bias. Besides, to reduce the impact of outliers, we use the Mean Absolute Error (MAE) as a substitute for MSE, which helps to minimize the amplification of errors caused by quantizing outliers.
The loss of quantizing the $i$-th block (${\rm Qblock}_i$) can be obtained by computing Mean Absolute Error (MAE) between quantized outputs $\hat{y}_{i+k-1}$ and full-precision outputs $y_{i+k-1}$,
where $k$ denotes the number of used blocks, $x_i$ denotes the raw inputs for the original $i$-th decoder ${\rm block_i}$. 
The parameters are optimized using the following equation:
\vspace{-3mm}
\begin{equation}
\label{eq:optimize}
\vspace{-4mm}
\operatorname*{argmin}_{\gamma,\beta,s,\delta}  \text{MAE}(\hat{y}_{i+k-1}, y_{i+k-1}).
\end{equation}
\vspace{-3mm}

%% file: sec/3_experiments.tex

\begin{table*}[!tbp]
\centering
\resizebox{\linewidth}{!}{
\begin{tabular}{lllccccc}
\hline
\textbf{Model} & \textbf{Method} & \textbf{Setting} & \textbf{Longtext avg$\uparrow$} & \textbf{Zero-shot avg$\uparrow$} & \textbf{WikiText2 ppl$\downarrow$} & \textbf{PTB ppl$\downarrow$} & \textbf{~C4 ppl$\downarrow$~} \\
\hline
\multirow{6}{*}{LLaMA-2-7B}
 & - & FP16 & 34.69 & 58.30\% & 5.47 & 37.91 & 7.26\\
 & GPTQ & W4 & 34.71 & 57.74\% & 5.62 & 241.52 & 7.46\\
 & OmniQuant & W4A4 & 11.16 & 45.44\% & 14.34 & 10304.68 & 20.18\\
 & OmniQuant$^\dagger$ & W4KV4 & 31.34 & 56.75\% & 6.09 & 64.39 & 8.98\\
 \rowcolor{gray!25}
 \cellcolor{white} & \textbf{WKVQuant} & W4KV4 & 34.48 & 58.38\% & 5.64 & 38.85 & 7.49\\
\hline
\multirow{6}{*}{LLaMA-2-13B}
 & - & FP16 & 34.12 & 61.32\% & 4.88 & 50.93 & 6.72\\
 & GPTQ & W4 & 34.06 & 60.55\% & 4.99 & 50.25 & 6.84\\
 & OmniQuant & W4A4 & 16.35 & 47.46\% & 12.39 & 263.60 & 17.51\\
 & OmniQuant$^\dagger$ & W4KV4 & 31.72 & 59.82\% & 5.18 & 55.95 & 7.30\\
 \rowcolor{gray!25}
 \cellcolor{white} & \textbf{WKVQuant} & W4KV4 & 32.52 & 60.34\% & 5.00 & 52.36 & 6.89\\
\hline
\multirow{6}{*}{LLaMA-7B}
 & - & FP16 & 34.80 & 57.68\% & 5.67 & 41.15 & 7.34\\
 & GPTQ & W4 & 33.54 & 57.30\% & 5.83 & 43.70 & 7.51\\
 & OmniQuant & W4A4 & 7.83 & 46.53\% & 11.58 & 231.46 & 16.19\\
 & OmniQuant$^\dagger$ & W4KV4 & 28.80 & 56.84\% & 6.25 & 51.96 & 8.17\\
 \rowcolor{gray!25}
 \cellcolor{white} & \textbf{WKVQuant} & W4KV4 & 33.86 & 57.91\% & 5.80 & 44.02 & 7.54\\
\hline
\multirow{6}{*}{LLaMA-13B}
 & - & FP16 & 36.02 & 60.60\% & 5.09 & 28.09 & 6.79\\
 & GPTQ & W4 & 35.14 & 60.10\% & 5.19 & 29.25 & 6.90\\
 & OmniQuant & W4A4 & 11.55 & 43.56\% & 11.18 & 115.38 & 16.37\\
 & OmniQuant$^\dagger$ & W4KV4 & 28.79 & 58.92\% & 5.64 & 32.78 & 7.86\\
 \rowcolor{gray!25}
 \cellcolor{white} & \textbf{WKVQuant} & W4KV4 & 35.50 & 60.44\% & 5.21 & 27.74 & 6.93\\
\hline
\end{tabular}
}
\caption{\label{table:main_result}
The overall experimental results. Our results are shown in gray line. 
}
\end{table*}

\begin{table*}[!thb]
\centering
\resizebox{\linewidth}{!}{
\begin{tabular}{lllcccccc}
\hline
\textbf{Model} & \textbf{Method} & \textbf{Setting} & \textbf{Qasper$\uparrow$} & \textbf{ 2WikiMultihopQA$\uparrow$} & \textbf{HotpotQA$\uparrow$} & \textbf{TriviaQA$\uparrow$} & \textbf{LCC$\uparrow$} & \textbf{Longtext avg$\uparrow$}\\
\hline
\multirow{6}{*}{LLaMA-2-7B}
  & - & FP16 & 7.81 & 7.53 & 8.44 & 84.21 & 65.49 & 34.69 \\
  & GPTQ & W4 & 7.68 & 9.35 & 8.55 & 84.88 & 63.12 & 34.71 \\
  & OmniQuant & W4A4 & 4.47 & 5.51 & 3.94 & 18.55 & 23.35 & 11.16 \\
  & OmniQuant$^\dagger$ & W4KV4 & 6.18 & 8.69 & 7.07 & 77.81 & 56.96 & 31.34 \\
  \rowcolor{gray!25}
  \cellcolor{white} & \textbf{WKVQuant} & W4KV4 & 7.57 & 9.64 & 8.31 & 84.11 & 62.79 & 34.48 \\
\hline
\multirow{6}{*}{LLaMA-7B}
& - & FP16 & 7.19 & 9.60 & 9.70 & 83.49 & 64.04 & 34.80 \\
& GPTQ & W4 & 6.31 & 9.91 & 9.36 & 79.51 & 62.61 & 33.54 \\
& OmniQuant & W4A4 & 2.77 & 4.57 & 3.54 & 13.19 & 15.11 & 7.83 \\
& OmniQuant$^\dagger$ & W4KV4 & 4.53 & 8.01 & 7.51 & 66.23 & 57.74 & 28.80 \\
\rowcolor{gray!25}
\cellcolor{white} & \textbf{WKVQuant} & W4KV4 & 6.75 & 10.23 & 9.18 & 81.21 & 61.93 & 33.86 \\
\hline
\end{tabular}
}
\caption{\label{table:All LongBench score results}
Longtext scores. Results of LLaMA-2-13B and LLaMA-13B can be found in \ref{sec:Longtext Scores of LLaMA-2-13B and LLaMA-13B models}.
}
\vspace{-4mm}
\end{table*}
\section{Experiments}
\vspace{-1mm}
In this section, we introduce the detailed experimental settings as well as demonstrate the effectiveness of the proposed \emph{WKVQuant}.

\subsection{Experimental Settings}
\paragraph{Models.} We evaluate our \emph{WKVQuant} on LLaMA~\cite{touvron2023llama} and LLaMA-2~\cite{touvron2023llama2} models (i.e., LLaMA-2-7B, LLaMA-2-13B, LLaMA-7B, and LLaMA-13B). 

\vspace{-1mm}
\paragraph{Baselines.} We perform our \emph{WKVQuant} in W4KV4 (quantizing weights to 4 bit and keys/values to 4 bit) setting. We also display results on OmniQuant~\cite{2023omniquant} in W4A4 (quantizing weights to 4 bit and activations to 4 bit) setting and on GPTQ~\cite{frantar2022gptq} in W4 setting. Since there is no work for weight-KV/cache quantization before, we combine the scaling parameter for weights in W4 OmniQuant and the scaling parameter between query and key states in W4A4 OmniQuant to formalize so-called OmniQuant$^\dagger$ in W4KV4 setting for comparison. The weight quantization in all methods are performed with group size of 128.

\vspace{-1mm}
\paragraph{Calibration.} The calibration dataset contains 128 randomly selected 2048-token segments from WikiText2~\cite{merity2016pointer}. The scaling factor $s$ and shifting factor $\delta$ are initialized by computing the maximum absolute value and the mean value of corresponding feature representation across channels in the calibration dataset, respectively. The group size for KV cache quantization is set as 128. The learnable clipping factors $\gamma$ and $\beta$ are initialized as in OmniQuant. 
To optimize the learnable parameters, we utilize AdamW optimizer with zero weight decay. The learning rate for two-dimensional quantization (2D-Quantization) and learnable weight clipping is set as 5e\textminus4 and 1e\textminus2, respectively. We utilize Cross-block Reconstruction Regularization (CRR) for optimizing and the hyper-parameter $k$ is fixed to 5 (see \ref{sec:Find k of CRR}). The optimization process is facilitated on a single Nvidia A100 GPU, utilizing a batch size of 1 over 5 epochs. Since the KV cache is not used in our calibration dataset, we deactivate past only quantization (POQ) while training and activate POQ after training.

\vspace{-1mm}
\paragraph{Evaluated tasks and datasets.}To eliminate the effectiveness of quantized LLMs under long input sequence circumstances, we perform experiments by using LongBench~\cite{bai2023longbench} framework on various datasets, i.e., Qasper~\cite{dasigi2021dataset}, 2WikiMultihopQA~\cite{ho2020constructing}, HotpotQA~\cite{yang2018hotpotqa}, TriviaQA~\cite{joshi2017triviaqa}, LCC~\cite{guo2023longcoder}, and MultiFieldQA-en~\cite{bai2023longbench}. The corresponding results are shown as Longtext score. We display the detailed information of the used datasets, which can be found in \ref{sec:Detailed information of the used LongBench datasets}.
Following previous works, we also evaluate them on five zero-shot common sense reasoning tasks, i.e., PIQA~\cite{tata2003piqa}, ARC-Challenge~\cite{clark2018think}, HellaSwag~\cite{zellers2019hellaswag}, and WinoGrande~\cite{sakaguchi2021winogrande}. Moreover, we basically evaluate the quantized models’ perplexity (ppl) scores on WikiText2~\cite{merity2016pointer}, Pen Treebank (PTB)~\cite{marcus1994penn} and C4~\cite{raffel2020exploring}.

\subsection{Overall Results}
As shown in Table \ref{table:main_result}, we conduct various experiments to evaluate our proposed \emph{WKVQuant}. In general, we can find that methods in W4KV4 setting have obvious improvement compared to the W4A4 setting, highlighting the value of quantizing only the KV cache within the range of activations. As shown in Table \ref{table:All LongBench score results}, \emph{WKVQuant} outperforms OmniQuant$^\dagger$ in terms of average performance on the Longtext evaluation datasets, indicating the superior accuracy of our method for weigh-KV cache quantization. Notably, the evaluation processes of zero-shot accuracy and perplexity do not utilize KV cache, thus the results of \emph{WKVQuant} on these tasks are turned to be under the W4 setting because of POQ. In this case, \emph{WKVQuant} exhibits comparable performance with GPTQ, demonstrating the feasibility of co-optimizing the parameters for weights and KV cache quantization. It is worth mentioning that \emph{WKVQuant} provides comparable performance with GPTQ in Longtext datasets, suggesting that it can effectively quantize huge memory consumption caused by KV cache based on weight-only quantization, without introducing significant accuracy degradation. For detailed zero-shot results, refer to \ref{sec:All zero-shot accuracy results}.

\vspace{-1mm}
\subsection{Ablation Study}
\vspace{-2mm}

\begin{table}[!tbp]
\centering
\resizebox{\linewidth}{!}{
\begin{tabular}{ll}
\hline
\textbf{Quantization Method} & \textbf{Evaluation Score$\uparrow$}\\
\hline
FP16 & 24.42\\
\hline
RTN & 2.3\\
+ LWC & 3.88 (1.58$\uparrow$)\\
+ 2D-Quant-channel & 19.03 (16.73$\uparrow$)\\
+ 2D-Quant-token & 19.37 (17.07$\uparrow$)\\
+ POQ & 16.11 (13.81$\uparrow$)\\
\hline
\end{tabular}
}
\caption{\label{table:Ablation study compared with RTN}
The accuracy gain by utilizing each technique based on quantized LLaMA-2-7B model with RTN method. 
}
\vspace{-2mm}
\end{table}


\begin{table}[!tbp]
\centering
\resizebox{\linewidth}{!}{
\begin{tabular}{ll}
\hline
\textbf{Quantization Method} & \textbf{Evaluation Score$\uparrow$}\\
\hline
FP16 & 24.42\\
\hline
WKVQuant & 25.29\\
- LWC & 24.08 (1.21$\downarrow$)\\
- 2D-Quant-channel & 22.74 (2.55$\downarrow$)\\
- 2D-Quant-token & 23.14 (2.15$\downarrow$)\\
- POQ & 19.95 (5.34$\downarrow$)\\
\hline
\end{tabular}
}
\caption{\label{table:Ablation study compared with WKVQuant}
The accuracy drop by deactivating each technique based on quantized LLaMA-2-7B model with \emph{WKVQuant} method. 
}
\vspace{-2mm}
\end{table}

\paragraph{Ablation study for POQ and 2D-Quantization.} To evaluate the effectiveness of the proposed POQ and 2D-Quantization, we first perform ablation study on LLaMA-2-7B and MultiFieldQA-en dataset with LongBench framework in W4KV4 setting considering that the methodology of POQ changes W4KV4 setting to W4 setting if KV cache is not used. We first apply different techniques used in \emph{WKVQuant},i.e., learnable weight clipping (LWC), static channel-wise smoothing (2D-Quant-channel), dynamic token-wise fine-grained quantization (2D-Quant-token), and POQ, to the RTN model separately to evaluate their contributions. Here, the RTN method refers to the vanilla round-to-nearest quantization. As shown in Table \ref{table:Ablation study compared with RTN}, each part of \emph{WKVQuant} has positive effects on model performance especially POQ and 2D-Quantization. We then deactivate those used techniques from complete \emph{WKVQuant} method to observe the accuracy change with their absence. Results in Table \ref{table:Ablation study compared with WKVQuant} shows that POQ holds paramount significance for performance, followed by 2D-Quantization. 


\begin{table}[!tbp]
\centering
\resizebox{\linewidth}{!}{
\begin{tabular}{lcc}
\hline
\textbf{Longtext Dataset} & \textbf{Block-wise MSE} & \textbf{CRR} \\
\hline
Qasper$\uparrow$ & 7.13 & 7.57 \\
2WikiMultihopQA$\uparrow$ & 9.41 & 9.64 \\
HotpotQA$\uparrow$ & 7.99 & 8.31 \\
TriviaQA$\uparrow$ & 84.60 & 84.11 \\
LCC$\uparrow$ & 61.24 & 62.79 \\
Loingtext avg$\uparrow$ & 34.07 & 34.48 \\
\hline
\end{tabular}
}
\caption{\label{table:Ablation study for Cross-block Reconstruction Regularization}
Ablation study of CRR. 
}
\end{table}

\paragraph{Ablation study for CRR.} Here, we ablate the effect of proposed Cross-block Reconstruction Regularization (CRR). As shown in Table \ref{table:Ablation study for Cross-block Reconstruction Regularization}, the proposed CRR is more effective than the commonly utilized block-wise MSE loss, which demonstrate the benefits
training parameters by computing a more proper loss from a more global perspective.

\subsection{Memory Consumption}
\begin{table}[!tbp]
\resizebox{\linewidth}{!}{
\begin{tabular}{@{}llcccc@{}}
\toprule
                               & Model       & FP   & W4   & W4KV4 & W4A4 \\ \midrule
\multirow{2}{*}{bs=1 len=2048} & LLaMA-2-7b  & 14.0 & 4.3  & 3.5   & 3.5  \\
                               & LLaMA-2-13b & 27.1 & 8.0  & 6.8   & 6.8  \\
\multirow{2}{*}{bs=1 len=9012} & LLaMA-2-7b  & 17.2 & 7.5  & 4.3   & 4.3  \\
                               & LLaMA-2-13b & 32.1 & 13.1 & 8.0   & 8.0  \\
\multirow{2}{*}{bs=16 len=2048} & LLaMA-2-7b  & 30.1 & 20.4 & 7.5   & 7.5  \\
                               & LLaMA-2-13b & 52.2 & 33.2 & 13.1  & 13.1 \\ \bottomrule
\end{tabular}
}
\caption{\label{table:memory_consumption}
The memory consumption (GB) of decoding using different quantization settings.
}
\vspace{-2mm}
\end{table}

We use LLMViewer~\cite{hahnyuanLLMViewer} to analyze memory consumption and show the results in Table~\ref{table:memory_consumption}. 
It can be observed that the weight-KV cache quantization has comparable memory savings with the weight-activation quantization, which means that \emph{WKVQuant} improves the accuracy with almost negligibly increased memory.

\section{Conclusion}
In this paper, we have identified the limitations of existing quantization approaches in balancing accuracy and efficiency for LLMs. To address these limitations, we propose \emph{WKVQuant}, a novel quantization methodology specifically designed for quantizing weights and the KV cache of LLMs. WKVQuant incorporates past-only quantization to optimize the improve the computation of attention. Additionally, we introduce two-dimensional quantization to handle the distribution of KV cache, along with a cross-block reconstruction loss for parameter optimization. Our proposed \emph{WKVQuant} provides a promising trade-off between accuracy and efficiency, making LLMs more practical for deployment in resource-constrained environments.

%% file: sec/4_appendix.tex
\newpage
\section{Appendix}\label{sec:appendix}

\subsection{Find the Best Number of Blocks for Cross-block Reconstruction Regularization}\label{sec:Find k of CRR}
We propose the Cross-block Reconstruction Regularization for our \emph{WKVQuant}, which computes Mean Absolute Error loss between the original and quantized outputs after several decoder blocks. To find out the best number of blocks, we perform ablation study designed for this hyper-parameter. Results are displayed in Table~\ref{table:Find k of CRR}. From the table we can observe that the best number for $k$ should be 5. So we fix it as 5 during the whole experiments.
\begin{table}[!htbp]
\centering
\begin{tabular}{cccc}
\hline
\textbf{k} & \textbf{WikiText2 ppl$\downarrow$} & \textbf{PTB ppl$\downarrow$} & \textbf{~C4 ppl$\downarrow$~}\\
\hline
    1 & 5.7094 & 42.9600 & 7.5700 \\
    2 & 5.6758 & 43.7015 & 7.5517 \\
    3 & 5.6677 & 46.0000 & 7.5498 \\
    4 & 5.6668 & 41.9152 & 7.5417 \\
    5 & 5.6622 & 39.7035 & 7.5413 \\
    6 & 5.6642 & 40.7976 & 7.5437 \\
\hline
\end{tabular}
\caption{\label{table:Find k of CRR}
Results of perplexity on WikiText2, PTB, and C4 dataset by applying different number of blocks used in Cross-block Reconstruction Regularization. It can be observed from the table that setting $k$ to 5 lead to the best results.
}
\end{table}

\subsection{Detailed Information of the Used LongBench Datasets}\label{sec:Detailed information of the used LongBench datasets}
To perform experiments on long texts (Longtexts), we utilize LongBench~\cite{bai2023longbench} framework to perfrom predictions on various datasets. Detailed information of the used datasets are shown in Table~\ref{table:LongBench datasets}. The first four datasets each have 200 samples. LCC dataset has 500 samples and MultiFieldQA-en dataset has 150 samples. 
\setcellgapes{4pt} 
\begin{table}[!htbp]
\centering
\resizebox{\linewidth}{!}{
\begin{tabular}{llcc}
\hline
\textbf{Dataset} & \textbf{Task Type} & \textbf{Metric} & \textbf{Length}\\
\hline
 Qasper & Single-doc QA & F1 & 3619\\
 2WikiMultihopQA & Multi-doc QA & F1 & 4887\\
 HotpotQA &Multi-doc QA & F1 & 9151\\
 TriviaQA & Few-Shot QA & F1 & 8209\\
 LCC & Code & Edit Sim & 1235\\
 MultiFieldQA-en & Single-doc & F1 & 4559\\
\hline
\end{tabular}
}
\caption{\label{table:LongBench datasets}
Description of the used LongBench datasets.
}
\end{table}

\subsection{Implementation Details}
To optimize the learnable parameters, we utilize AdamW optimizer with zero weight decay. The learning rate for two-dimensional quantization (2D-Quantization) and learnable weight clipping is set as 5e\textminus4 and 1e\textminus2, respectively. The optimization process is facilitated on a single Nvidia A100 GPU, utilizing a batch size of 1 over 5 epochs. Since the KV cache is not used in our calibration dataset, we deactivate past only quantization (POQ) while training and activate POQ after training.

\subsection{Zero-shot Accuracy}\label{sec:All zero-shot accuracy results}
See Table~\ref{table:All zero-shot accuracy results}. This table can be seen as an extension to Table~\ref{table:main_result} as we display the zero-shot accuracy results on all the used datasets here.

\begin{table*}[!tbp]
\centering
\resizebox{\linewidth}{!}{
\begin{tabular}{lllccccc}
\hline
\textbf{Model} & \textbf{Method} & \textbf{Setting} & \textbf{~~PIQA$\uparrow$~~} & \textbf{ARC-Challenge$\uparrow$} & \textbf{HellaSwag$\uparrow$} & \textbf{WinoGrande$\uparrow$} & \textbf{Zero-shot avg$\uparrow$}\\
\hline
\multirow{6}{*}{LLaMA-2-7B}
  & - & FP16 & 78.40\% & 39.84\% & 56.67\% & 67.24\% & 58.30\%\\
  & GPTQ & W4 & 78.34\% & 38.82\% & 56.06\% & 66.92\% & 57.74\%\\
  & OmniQuant & W4A4 & 66.15\% & 28.15\% & 42.04\% & 53.67\% & 45.45\%\\
  & OmniQuant$^\dagger$ & W4KV4 & 77.09\% & 38.99\% & 54.17\% & 65.43\% & 56.75\%\\
  \rowcolor{gray!25}
  \cellcolor{white} & WKVQuant & W4KV4 & 78.23\% & 40.78\% & 56.14\% & 67.48\% & 58.38\%\\
\hline
\multirow{6}{*}{LLaMA-2-13B}
& - & FP16 & 78.72\% & 45.56\% & 59.69\% & 69.69\% & 61.32\%\\
& GPTQ & W4 & 78.83\% & 43.60\% & 59.24\% & 68.58\% & 60.55\%\\
& OmniQuant & W4A4 & 67.02\% & 30.54\% & 44.83\% & 53.90\% & 47.46\%\\
& OmniQuant$^\dagger$ & W4KV4 & 78.07\% & 43.17\% & 58.22\% & 68.74\% & 59.82\%\\
\rowcolor{gray!25}
\cellcolor{white} & WKVQuant & W4KV4 & 78.12\% & 43.94\% & 58.98\% & 68.75\% & 60.34\%\\
\hline
\multirow{6}{*}{LLaMA-7B}
& - & FP16 & 78.40\% & 38.22\% & 56.42\% & 66.92\% & 57.68\%\\
& GPTQ & W4 & 78.18\% & 37.62\% & 56.11\% & 65.98\% & 57.30\%\\
& OmniQuant & W4A4 & 66.97\% & 29.01\% & 43.63\% & 53.35\% & 46.53\%\\
& OmniQuant$^\dagger$ & W4KV4 & 77.63\% & 37.88\% & 55.02\% & 65.35\% & 56.84\%\\
\rowcolor{gray!25}
\cellcolor{white} & WKVQuant & W4KV4 & 78.23\% & 39.76\% & 55.75\% & 65.58\% & 57.91\%\\
\hline
\multirow{6}{*}{LLaMA-13B}
  & - & FP16 & 78.78\% & 43.94\% & 59.10\% & 70.08\% & 60.61\%\\
  & GPTQ & W4 & 78.29\% & 43.51\% & 58.50\% & 69.13\% & 60.10\%\\
  & OmniQuant & W4A4 & 64.25\% & 24.57\% & 41.88\% & 52.09\% & 43.56\%\\
  & OmniQuant$^\dagger$ & W4KV4 & 78.45\% & 40.95\% & 57.37\% & 70.32\% & 58.92\%\\
  \rowcolor{gray!25}
  \cellcolor{white} & WKVQuant & W4KV4 & 79.71\% & 43.00\% & 58.62\% & 71.19\% & 60.44\%\\
\hline
\end{tabular}
}
\caption{\label{table:All zero-shot accuracy results}
Zero-shot accuracy of all methods on PIQA, ARC-Challenge, HellaSwag, and WinoGrande dataset.
}
\end{table*}

\begin{table*}[!tbp]
\centering
\resizebox{\linewidth}{!}{
\begin{tabular}{lllcccccc}
\hline
\textbf{Model} & \textbf{Method} & \textbf{Setting} & \textbf{Qasper$\uparrow$} & \textbf{ 2WikiMultihopQA$\uparrow$} & \textbf{HotpotQA$\uparrow$} & \textbf{TriviaQA$\uparrow$} & \textbf{LCC$\uparrow$} & \textbf{Longtext avg$\uparrow$}\\
\hline
\multirow{6}{*}{LLaMA-2-13B}
  & - & FP16 & 6.55 & 8.32 & 8.86 & 84.97 & 61.92 & 34.12 \\
  & GPTQ & W4 & 7.02 & 8.72 & 8.66 & 85.64 & 60.28 & 34.06 \\
  & OmniQuant & W4A4 & 4.71 & 5.41 & 4.70 & 28.96 & 37.98 & 16.35 \\
  & OmniQuant$^\dagger$ & W4KV4 & 6.45 & 8.79 & 8.09 & 81.58 & 53.69 & 31.72 \\
  \rowcolor{gray!25}
  \cellcolor{white} & \textbf{WKVQuant} & W4KV4 & 6.00 & 7.73 & 8.26 & 82.15 & 58.5 & 32.52 \\
\hline
\multirow{6}{*}{LLaMA-13B}
  & - & FP16 & 8.16 & 9.88 & 9.69 & 85.50 & 66.87 & 36.02 \\
  & GPTQ & W4 & 6.97 & 10.00 & 9.54 & 84.05 & 65.16 & 35.14 \\
  & OmniQuant & W4A4 & 3.75 & 4.84 & 4.44 & 21.09 & 23.66 & 11.55 \\
  & OmniQuant$^\dagger$ & W4KV4 & 5.35 & 8.31 & 8.33 & 66.14 & 55.84 & 28.79 \\
  \rowcolor{gray!25}
  \cellcolor{white} & \textbf{WKVQuant} & W4KV4 & 6.71 & 9.47 & 9.04 & 86.60 & 65.69 & 35.50 \\
\hline
\end{tabular}
}
\caption{\label{table:additional LongBench scores}
Longtext scores. Our results are shown in gray line. 
}
\vspace{-4mm}
\end{table*}

\subsection{Longtext Scores of LLaMA-2-13B and LLaMA-13B models}\label{sec:Longtext Scores of LLaMA-2-13B and LLaMA-13B models}
See Tabel~\ref{table:additional LongBench scores}. This table is a supplement to Table~\ref{table:All LongBench score results} as we display Longtext scores of LLaMA-2-13B and LLaMA-13B models here.